\documentclass[lettersize,journal]{IEEEtran}
\usepackage{amsmath,amsfonts}
\usepackage{algorithmic}
\usepackage{algorithm}
\usepackage{array}
\usepackage[caption=false,font=normalsize,labelfont=sf,textfont=sf]{subfig}
\usepackage{textcomp}
\usepackage{stfloats}
\usepackage{url}
\usepackage{verbatim}
\usepackage{graphicx}
\usepackage{cite}
\usepackage{mdframed}
\usepackage[svgnames,table]{xcolor}  

\begin{document}

\title{STAA: Spatio-Temporal Attention Attribution for Real-Time Interpreting Transformer-based Video Models}


\author{Zerui Wang, \IEEEmembership{Student Member, IEEE}, and Yan Liu, \IEEEmembership{Member, IEEE}}


\markboth{IEEE TRANSACTIONS ON CLOUD COMPUTING,~VOL.~XX, NO.~XX, MONTH~2024}%
{Author \MakeLowercase{\textit{et al.}}: Paper Title}


\maketitle

\begin{abstract}
        Transformer-based models have achieved state-of-the-art performance in various computer vision tasks, including image and video analysis. However, Transformer's complex architecture and black-box nature pose challenges for explainability, a crucial aspect for real-world applications and scientific inquiry. 
        Current Explainable AI (XAI) methods can only provide one-dimensional feature importance, either spatial or temporal explanation, with significant computational complexity.
        This paper introduces STAA (Spatio-Temporal Attention Attribution), an XAI method for interpreting video Transformer models. Differ from traditional methods that separately apply image XAI techniques for spatial features or segment contribution analysis for temporal aspects, STAA offers both spatial and temporal information simultaneously from attention values in Transformers.
        The study utilizes the Kinetics-400 dataset, a benchmark collection of 400 human action classes used for action recognition research. We introduce metrics to quantify explanations.
        We also apply optimization to enhance STAA's raw output. By implementing dynamic thresholding and attention focusing mechanisms, we improve the signal-to-noise ratio in our explanations, resulting in more precise visualizations and better evaluation results. 
        In terms of computational overhead, our method requires less than 3\% of the computational resources of traditional XAI methods, making it suitable for real-time video XAI analysis applications.
        STAA contributes to the growing field of XAI by offering a method for researchers and practitioners to analyze Transformer models. 
\end{abstract}


\section{Introduction}

\IEEEPARstart{V}{ideo} understanding has emerged as a cornerstone of modern artificial intelligence, revolutionizing critical applications from autonomous driving safety systems \cite{Yuan2020TemporalChannelTF} to precision medical diagnostics \cite{Liu2023LoViTLV}. The introduction of Transformer architectures, exemplified by TimeSformer \cite{TimeSformer} and ViViT \cite{arnab2021vivit}, represents a paradigm shift in video analysis, consistently outperforming traditional convolutional neural networks (CNNs) \cite{kay2017kinetics, carreira2017quo} across diverse benchmarks. However, this leap in performance introduces new challenges: as model architectures grow more sophisticated, their decision-making processes become increasingly opaque, necessitating advanced explainable AI (XAI) approaches \cite{Nourani2020DontEW} to ensure accountability and trust in critical applications.


Video understanding presents unique challenges compared to static image analysis, primarily due to its inherent spatio-temporal complexity. While spatial features capture frame-level information such as object configurations and contextual relationships, temporal features encode crucial dynamic patterns—motion trajectories, state transitions, and temporal dependencies \cite{Wang2016Temporal}. The significance of temporal modeling is evidenced by breakthroughs such as Temporal Segment Networks (TSN) \cite{Wang2016Temporal}, which demonstrate how precise temporal feature analysis enables the discrimination of visually similar but temporally distinct actions. This spatio-temporal interplay forms the foundation of advanced video understanding tasks, where the precise sequencing and timing of spatial transformations define complex actions \cite{carreira2017quo}. Understanding and explaining these intricate spatio-temporal relationships becomes crucial as video AI systems increasingly influence high-stakes decisions in autonomous systems \cite{Yuan2020TemporalChannelTF} and healthcare \cite{Liu2023LoViTLV}.


While significant advances have been made in developing XAI methods for computer vision \cite{wang2024sse}, particularly in CNN-based image analysis \cite{selvaraju2017grad}, extending these approaches to Transformer-based video models presents unprecedented challenges. These challenges manifest in two critical dimensions: algorithmic complexity and computational efficiency, both of which fundamentally impact the practical deployment of explainable video AI systems.


\textbf{Algorithmic Challenges:}
The distinctive architecture of Transformers, centered on self-attention mechanisms, represents a fundamental departure from traditional CNNs, rendering many established XAI methods ineffective or theoretically incompatible \cite{selvaraju2017grad,chefer2021transformer}. 
Besides, architectural divergence is compounded by the inherent complexity of video data, where temporal dynamics create additional complexity beyond what static image-based XAI methods can address \cite{zhu2020comprehensive}. Current approaches often artificially dichotomize spatial and temporal analysis, producing fragmented explanations that fail to capture the holistic nature of video understanding. This limitation creates a critical gap in our ability to explain how models interpret and process the dynamic interplay between spatial and temporal features in video content.


\textbf{Computational Efficiency Challenges:}
The computational demands of existing post-hoc explanation methods present a significant barrier to real-world deployment. These methods typically require multiple model inference passes, creating substantial computational overhead for video analysis tasks. This inefficiency becomes particularly acute in scenarios demanding real-time explanations, such as autonomous systems or live video monitoring applications \cite{Milliseconds}. The computational burden is especially problematic for long video sequences, where the resource requirements can grow exponentially \cite{lundberg2017unified,xaiprocess} due to the algorithm of post-hoc methods.


These challenges underscore the need for XAI methods to adapt to transformer architectures, provide spatio-temporal explanations, and achieve high computational efficiency. We pose the following research questions:

\textit{\textbf{RQ1: Can we develop a novel XAI method for video Transformer models that simultaneously captures spatial and temporal feature explanations?}} 

\textit{\textbf{RQ2: Can our proposed XAI method improve faithfulness, monotonicity, and computational efficiency compared to traditional model-agnostic methods in video analysis tasks?}}

\textit{\textbf{RQ3: Can we design and implement a cloud-based XAI service architecture that enables real-time explanations for edge-side video?}}


To the best of our knowledge, there are several critical gaps in the current video AI explainability:
(1) Despite the widespread deployment of Transformer-based video \cite{TimeSformer,vivit,wang2024sse}, there remains an absence of methods for video Transformer architectures \cite{selva2023video}. 
(2) Current approaches provide a single type of feature explanation, failing to simultaneously capture both spatial and temporal features in video models.
(3) The absence of real-time capabilities for video analysis creates a significant barrier for edge computing applications \cite{edgecloudecosystem}. Traditional post-hoc approaches fail to address the computational demands of video analysis. 
These gaps collectively impede practitioners from understanding and trusting video AI systems, highlighting the urgent need for novel XAI approaches that can address both the theoretical and practical challenges of explaining video Transformer models.

To address these critical gaps, we propose STAA (Spatio-Temporal Attention Attribution), a novel XAI method for video Transformer models. STAA achieves three key advantages: (1) simultaneous capture of spatial and temporal feature importance through a single forward pass, (2) improved explanation faithfulness through direct alignment with the model's decision-making process, and (3) significantly reduced computational overhead by utilizing the model's internal attention mechanisms rather than requiring multiple inference passes.
In this paper, the main contributions are:

\textbf{Novel XAI Architecture:} STAA fundamentally addresses the architectural mismatch between current XAI methods and video Transformer models.
    
\textbf{Empirical Validation:} Rigorous evaluation of STAA against adapted versions of SHAP and LIME on the Kinetics-400 dataset, demonstrating significant improvements in both explanation quality and computational efficiency. 
    
\textbf{Real-Time XAI Framework:} Design and implementation of a cloud-based service architecture that enables real-time video explanations with sub-100ms latency, making STAA practical for edge computing applications. 



    
    

The remainder of this paper is organized as follows: Section II reviews related work in video Transformers and current XAI approaches, highlighting limitations in explaining video models. Section III presents our STAA method and adaptations of SHAP and LIME for video analysis. Section IV details our experimental setup using the Kinetics-400 dataset and TimeSformer architecture, introducing metrics for assessing video XAI quality. Section V demonstrates STAA's performance improvements in both explanation quality and computational efficiency, including our real-time implementation results. Section VI concludes with key findings and future directions.


\section{Related Work}

\subsection{Transformer Models for Video Analysis}

The success of Transformer architectures in natural language processing \cite{vaswani2017attention} has inspired researchers to adopt them to computer vision tasks \cite{vit}. Differ from traditional CNNs, Transformer-based models can capture long-range dependencies and have shown remarkable performance in various video understanding tasks \cite{kay2017kinetics}.
TimeSformer by Meta \cite{TimeSformer} pioneered the application of pure Transformer architectures to video classification by treating different frame patches and time steps as tokens. This approach demonstrated competitive performance on standard benchmarks while being computationally efficient. Building upon this work, ViViT by Google \cite{vivit} proposed a tubelet embedding scheme, further slightly improving video classification accuracy. Moreover, another work VideoSwin by Microsoft \cite{videoswin} adapted the hierarchical Swin Transformer \cite{swin} for video understanding tasks. 

These models consistently outperformed CNN-based approaches in their paper. 
However, the previous XAI methods for generating saliency maps are not applicable to these Transformer-based video analysis models. 
While methods such as Grad-CAM \cite{selvaraju2017grad} have been effective in visualizing important regions in CNN-based models, they do not trivially extend to the self-attention mechanisms and token-based representations employed by Transformers.
This necessitates the development of novel XAI techniques specifically for Transformer architectures in video understanding tasks.
Recent work has begun to develop XAI methods for Transformer models. For instance, the recent work \cite{wang2024sse} proposed XAI pipelines for multiple Transformer models in image classification. Adapting and extending such approaches to video analysis represents a promising direction for research.

\subsection{Explainable AI in Computer Vision}

XAI has gained significant attention in recent years due to the increasing complexity of deep learning models and the need for transparency in AI systems \cite{xaiprocess}. Various XAI techniques have been developed and applied in the comprehensive XAI service frameworks \cite{Wang2024tcc}. SHAP (SHapley Additive exPlanations) \cite{lundberg2017unified} provides feature attribution based on coalitional game theory. LIME (Local Interpretable Model-agnostic Explanations) \cite{ribeiro2016should} offers another approach by approximating the behavior of complex models with interpretable surrogate models in the vicinity of a specific input, providing insights into local decision boundaries.

Despite the significant progress in XAI techniques for image analysis, such as SHAP \cite{lundberg2017unified}, LIME \cite{ribeiro2016should} and GradCAM \cite{selvaraju2017grad}. These methods face substantial challenges when applied to video models. The temporal feature inherent in video data poses an additional dimension for most existing XAI methods, which are primarily designed for static images and struggle to capture the temporal feature of video content. Furthermore, the architectural differences between Transformers and traditional CNNs create a mismatch, as many XAI techniques rely on gradient maps \cite{selvaraju2017grad}, rendering them incompatible with Transformer models.

The computational demands of video analysis exacerbate these issues, as post-hoc explanation methods often require multiple inferencing runs, which is prohibitively expensive for deep models and long video sequences. 
The algorithm inefficiency limits their practical applicability in real-world scenarios where timely explanations are crucial \cite{edgecloudecosystem}. 
The current methods also fail to provide unified spatio-temporal explanations  \cite{lundberg2017unified,ribeiro2016should}. This approach fails to capture the intricate interactions between space and time that are fundamental to video data, resulting in incomplete or misleading interpretations of model behavior \cite{stergiou2021analyzing}.

These limitations underscore the pressing need for XAI approaches specifically designed to address the challenges posed by video Transformer models. Such methods must handle both spatial and temporal features while maintaining computational efficiency.

\subsection{Challenges in XAI for Video Transformer Models}

While significant progress has been made in developing XAI techniques for image analysis, the interpretability of video models presents unique challenges due to the additional temporal dimension and the complexity of Spatio-Temporal reasoning \cite{stergiou2021analyzing}.
For CNN-based video models, Hiley et al. \cite{hiley2019discriminating} adapted autograd for 3D CNNs to generate class activation maps for action recognition models. Building on this work, Stergiou et al. \cite{stergiou2019class} proposed Class-Specific Saliency Tubes (CSST), which extend 2D saliency maps to the temporal domain, providing frame-wise visualizations of important regions for action recognition.
However, their technique routes are not applicable to Transformer-based video models due to their fundamentally different architectures. 
To the best of our knowledge, the field of XAI for video Transformers remains unexplored.
With the Transformer achieving the performance and being more commonly deployed, lacking XAI becomes particularly concerning given the increasing adoption of these models in applications \cite{DARPA}.
We seek to provide deeper insights into their decision-making processes and pave the way for more transparent and trustworthy video analysis systems.

\subsection{Requirements for Real-Time Video XAI}

Software engineers define a real-time response as that occurs within 100 milliseconds \cite{Milliseconds}. Another work \cite{VideoInpainting} also define a real-time video inpainting system as one that can process a video frame in the range of hundreds of milliseconds.  We refer to this requirement for latency in real-time video analysis. Considering our STAA method reduce the computation time of XAI for a video clip from minutes-level to 0.16 seconds, it is possible to apply STAA in real-time video XAI analysis. 

Real-time video stream model AI has diverse applications across critical industries. In autonomous driving systems \cite{Yuan2020TemporalChannelTF}, real-time video analysis enables immediate detection of road hazards, pedestrians, and changing traffic conditions, allowing for split-second decision-making essential for passenger safety. In medical diagnostics \cite{Liu2023LoViTLV}, real-time video AI assists surgeons during procedures and enables rapid analysis of endoscopic footage for early disease detection. In security applications \cite{Saikrishnan2023AutomatedOD}, these systems continuously monitor surveillance feeds to detect anomalies and identify potential threats, providing immediate alerts to security personnel. These high-stakes applications underscore the importance of immediate and interpretable explanations of AI decisions.

\section{Spatio-Temporal XAI Methods for Video Transformer Models}

In this section, we study three approaches for explaining video Transformer models:
(1) SHAP-based temporal analysis: Identifies important temporal segments across video frames. (2) LIME-based spatial analysis: Generates spatial explanations for individual video frames. (3) STAA: Our proposed method that extracts information from Transformer attention mechanisms for spatio-temporal explanations.

\subsection{SHAP-based Temporal Feature Attribution}

The SHAP (SHapley Additive exPlanations) \cite{lundberg2017unified} analysis aims to identify the importance of different temporal segments in a video for the model's decision-making process. This method provides insights into which moments in a video are influential for classification.

\subsubsection{SHAP Algorithm Adaptation for Video Analysis}

SHAP applies concepts from coalitional game theory, particularly Shapley values, to feature attribution. In our video analysis, we map these concepts as follows: The temporal segments of the video serve as the ``players" in the game-theoretic sense. The ``coalitions" are formed by various subsets of these temporal segments. The ``characteristic function" is represented by the model's prediction for any given subset of segments. 
The SHAP value for a segment quantifies its average marginal contribution to the model's prediction across all possible combinations of segments.

To analyze the temporal importance of different parts of a video, we segment the video stream into equal-length temporal segments. SHAP considers all possible combinations of these segments. Uniform segmentation ensures that each segment represents an equal duration of the video, providing a fair basis for this combinatorial analysis. 
Practitioners can increase the number of segments to achieve a more fine-grained analysis or decrease it for a more coarse-grained XAI analysis.

\begin{algorithm}
    \caption{SHAP-based Temporal Feature Attribution}
    \label{alg:shap_enhanced}
    \textbf{Input:}  $D$, $N$\\
    \textbf{Output:} $M_t$
    
    \begin{algorithmic}
    \STATE $s_i = \left\{f_t \mid t \in \left[\frac{(i-1)D}{N}, \frac{iD}{N}\right)\right\}, \quad i = 1, 2, ..., N$
    \STATE Initialize $M_t = \{0\}^N$
    \FOR{$i = 1$ \TO $N$}
        \STATE $\phi_i = \sum_{S \subseteq N \setminus \{i\}} \frac{|S|!(N-|S|-1)!}{N!} [f_x(S \cup \{i\}) - f_x(S)]$ 
        \STATE // When using approximation: \\
        // $\phi_i = \frac{1}{K} \sum_{k=1}^K [f_x(z^k \cup \{i\}) - f_x(z^k)]$
        \STATE $M_t[i] = \phi_i$
    \ENDFOR
    \RETURN $M_t$
    \end{algorithmic}
  {\small \textbf{Notation:} $D$: Duration of video, $N$: Number of segments, $f_x$: Model prediction function, $M_t$: Temporal attribution, $K$: Number of Monte Carlo samples (for approximation),  $z^k$: Random subset of segments (for approximation)}
\end{algorithm}

\subsubsection{Temporal Importance Quantification}

Algorithm \ref{alg:shap_enhanced} presents our SHAP-based temporal feature attribution method \cite{lundberg2017unified}. The algorithm begins by segmenting the video into N equal-length segments. For each segment, it calculates a SHAP value to quantify its importance in the model's decision-making process.
Them we have two methods for calculating SHAP values: exact calculation and Monte Carlo approximation. The exact method considers all possible combinations of segments, providing the most accurate results but at a higher computational cost. The Monte Carlo approximation estimates SHAP values by sampling a specified number of random segment combinations, significantly reducing computation time while maintaining a good estimate of the true SHAP values.

\begin{figure}[h]
    \centering
    \includegraphics[width=1\linewidth]{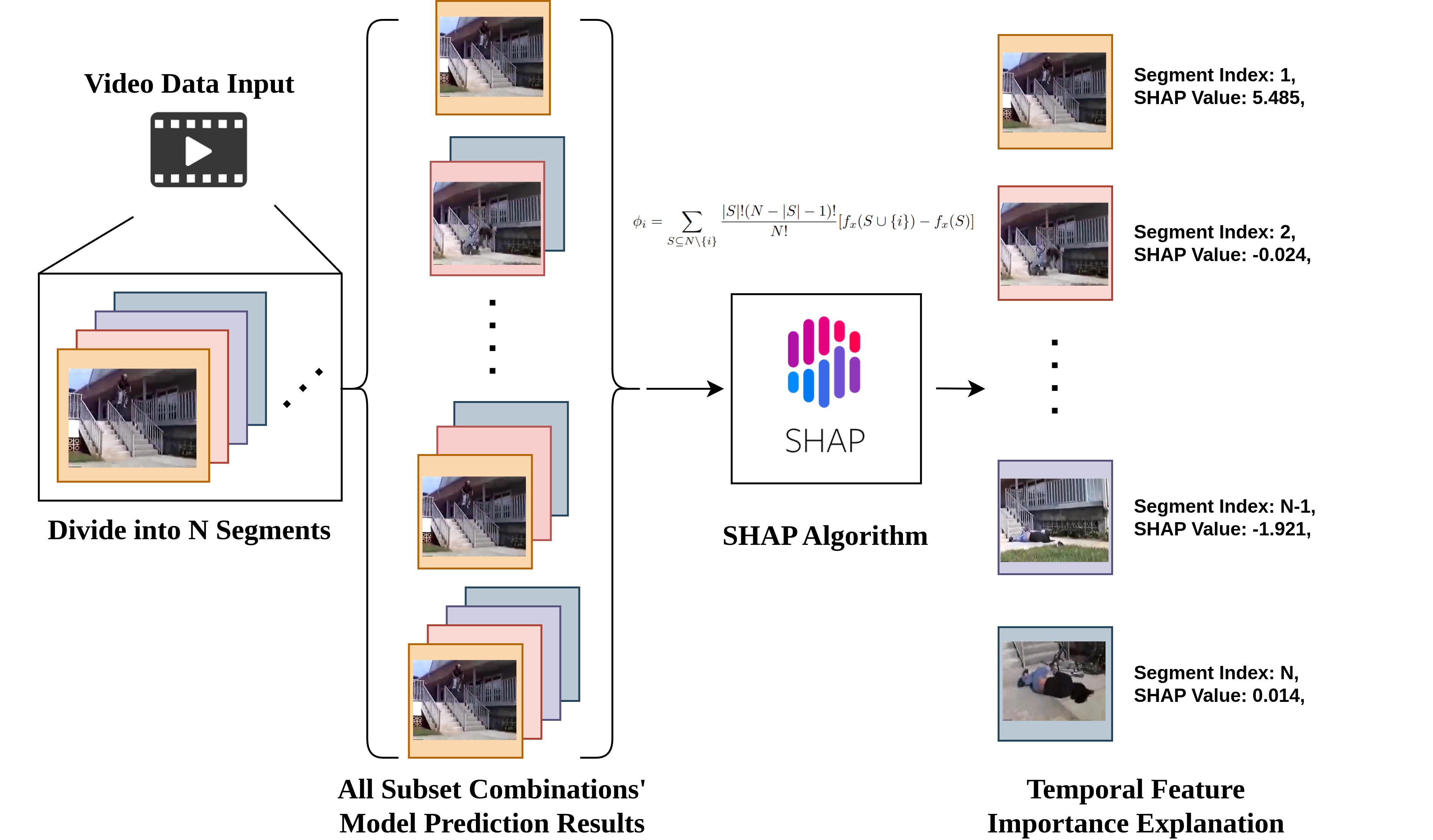}
    \caption{Flowchart of SHAP-based temporal feature attribution method}
    \label{fig:shap_flowchart}
    \end{figure}

To illustrate the process of SHAP-based temporal importance analysis, Figure \ref{fig:shap_flowchart} as a flowchart provides a clear understanding of the algorithm's workflow and how it identifies the most influential temporal regions in a video for the model's decision-making process.

\subsubsection{SHAP Computational Analysis}

Several factors affect the implementation of this method:
\texttt{Segment Size:} The choice of $N$ affects the granularity of the analysis. Smaller segments provide finer temporal resolution but increase computational cost. In our experiments, we set $N = 8$ for Kinetics-400, which offers a coarse temporal resolution. However, this already requires significant computational overhead. The computational complexity grows as $O(2^N)$ due to the need to evaluate all possible segment combinations.
\texttt{Approximation Strategy:} When computational resources are limited or for large-scale analyses, the Monte Carlo approximation can be employed. The number of Monte Carlo samples $K$ trades off between accuracy and computational efficiency. 
\texttt{Model Integration:} SHAP requires multiple inferences of the video model for different segment combinations. The computational requirements of SHAP are intrinsically linked to both the segmentation parameters and the underlying video AI model's complexity.

In summary, SHAP can provide temporal feature attribution through the scores of temporal segments. However, it has several limitations.
First, it focuses solely on one feature aspect, according to the segments. Second, the granularity of the analysis is limited by the number of segments $N$. A small $N$ provides only coarse resolution. 
The cost increases as $O(2^N)$ with the number of segments, making it challenging to analyze videos at high temporal resolutions or to process large datasets efficiently. While the Monte Carlo approximation mitigates this to some extent, it introduces a trade-off between accuracy and computational efficiency.



\subsection{LIME-based Spatial Feature Attribution}

LIME (Local Interpretable Model-agnostic Explanations) \cite{ribeiro2016should} is employed for spatial analysis of individual video frames. We adapt LIME with the Vision Transformer (ViT) model \cite{vit} trained on the Kinetics-400 dataset, enabling it to operate on individual frames while maintaining the model structure in Transformer.

\subsubsection{LIME Algorithm Adaptation for Video Analysis}

LIME is grounded in the principle of local approximation. It simplifies complex models by approximating them with interpretable surrogates in the vicinity of specific inputs. For video analysis, we adapt LIME to operate on individual frames extracted at regular intervals from the video sequence. The choice of frame count $K$ involves a trade-off between computational cost and temporal resolution.

For each selected frame, LIME generates perturbed variants by applying random masks to different regions, creating a local neighborhood around the original frame. The interpretable surrogate takes the form of a linear model operating on binary features, representing the presence or absence of interpretable components within the frame. This approach enables LIME to provide localized explanations by highlighting regions that significantly influence the model's frame-level decisions.

\begin{algorithm}
    \caption{LIME-based Spatial Feature Attribution}
    \label{alg:lime}
    \textbf{Input:}  $D$, $K$, $N_p$\\
    \textbf{Output:} $M_s$
    \begin{algorithmic}
    \STATE $f_k = \mathbf{V}(k \cdot \frac{D}{K}), \quad k = 1, 2, ..., K$
    \STATE Initialize $M_s = \{0\}^K$
    \FOR{$k = 1$ \TO $K$}
        \STATE $P_k = \{p_{k1}, \ldots, p_{kN_p}\}$
        \STATE $y_{km} = \text{ViT}(p_{km}), \quad m = 1, \ldots, N_p$
        \STATE $g_k = \arg\min_{g \in G} \mathcal{L}(g, \text{ViT}, P_k) + \Omega(g)$
        \STATE $M_s[k] = |g_k|$
    \ENDFOR
    \RETURN $M_s$
    \end{algorithmic}
    \small{\textbf{Notation:} $D$: Video duration, $K$: Number of frames, $N_p$: Number of perturbations, $G$: Class of interpretable models, $\mathcal{L}$: Loss function, $\Omega$: Regularization term}
\end{algorithm}

\subsubsection{Spatial Importance Quantification}
Algorithm \ref{alg:lime} presents the LIME-based spatial feature attribution method for video analysis:
The algorithm begins by extracting K equally spaced frames from the input video. For each frame $f_k$, it generates a set of M perturbed samples $P_k$ by applying random masks. These perturbations help explore the model's behavior in the vicinity of the original frame.
Next, the algorithm computes ViT predictions $y_{km}$ for each perturbed sample. An interpretable model $g_k$ is then fitted to approximate the ViT's behavior locally. This process balances between fidelity to the ViT's predictions (represented by the loss function $\mathcal{L}$) and model simplicity (enforced by the regularization term $\Omega$).
The algorithm computes importance scores $M_s[k]$ based on the fitted interpretable model for each frame. These scores form a sequence of spatial attributions that can be used to generate map visualization for each frame, with warmer colors indicating areas of higher importance.

\begin{figure}[h]
    \centering
    \includegraphics[width=1\linewidth]{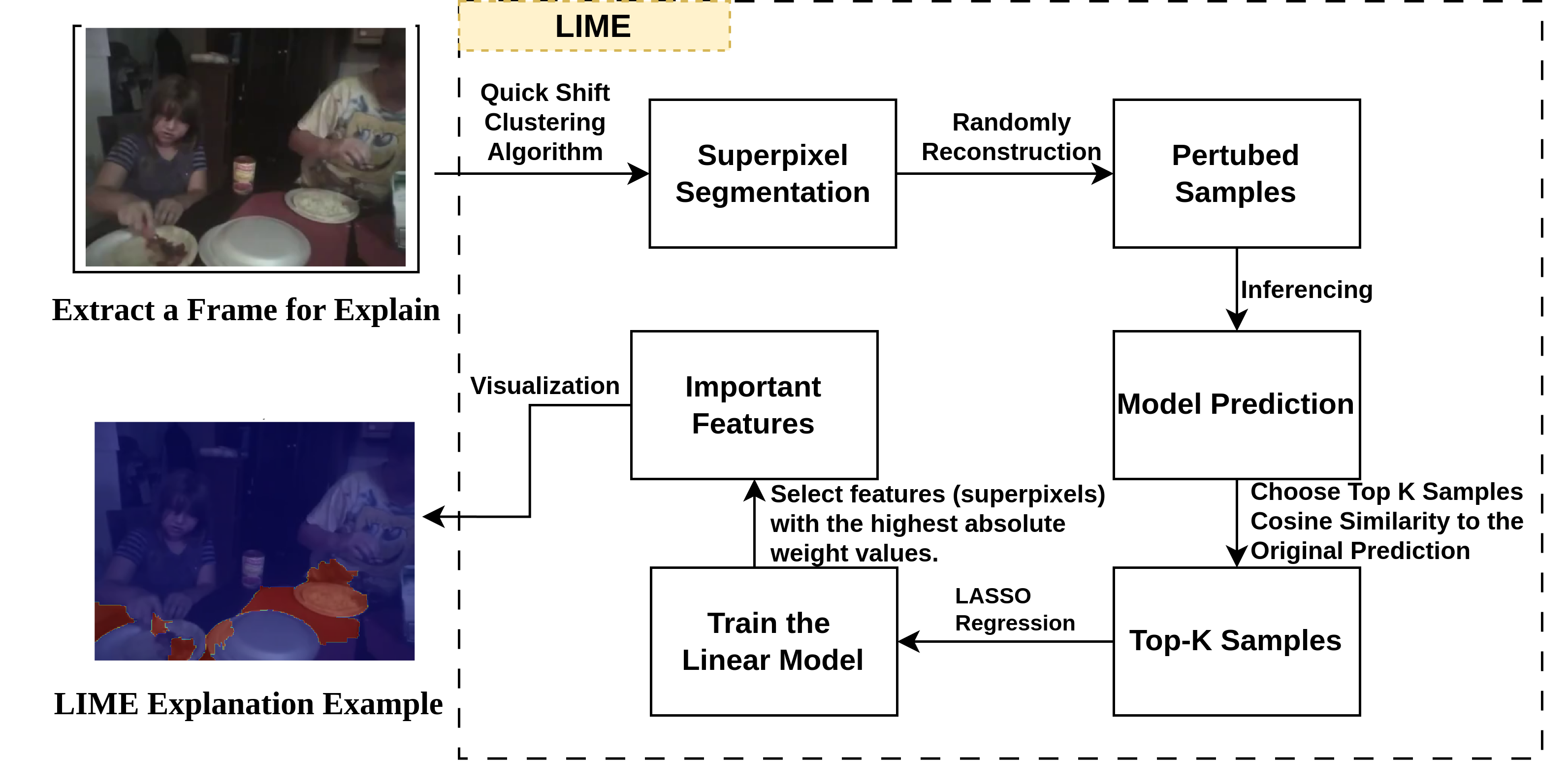}
    \caption{Flowchart of the LIME-based spatial feature attribution method}
    \label{fig:lime_flowchart}
\end{figure}

Figure \ref{fig:lime_flowchart} as a flowchart provides a visual representation of the LIME-based spatial feature attribution method.
By overlaying these results on the original frames, we can directly associate the importance scores with the visual content of the video, providing an interpretable representation of the model's focus areas. This approach allows us to apply LIME's image analysis capabilities while capturing relevant features for video analysis, providing insights into which spatial features are most influential in our video Transformer model's decision-making process.
This form of explanation is commonly used in vision XAI methods \cite{selvaraju2017grad,wang2024xaiport,wang2024sse} and provides a visual representation of the model's attention mechanisms.

\subsubsection{LIME Computational Analysis}

Several factors affect the implementation of LIME in our video analysis context:
\texttt{Number of Samples:} The number of perturbed samples $N_p$ significantly affects the fidelity of the local approximation. A larger $N_p$ generally leads to more accurate explanations but increases computational cost. In our experiments, we select $N_p = 1000$ for balanced accuracy and efficiency.
\texttt{Frame Selection:} The number and distribution of frames selected for analysis can affect the understanding of the video. 
\texttt{Perturbation Strategy:} The method of generating perturbations can influence the quality of explanations. We use the default setting in LIME \cite{ribeiro2016should}, which is random masking.

These factors should be considered when applying LIME in video XAI analysis.
LIME provides explanations that are local in nature and may ignore global patterns. Generating explanations for many frames can be computationally expensive, especially for high-resolution videos. LIME does not capture relationships between frames. 

\subsection{Spatio-Temporal Attention Attribution (STAA)}

\begin{figure*}[h]
\centering
\includegraphics[width=\textwidth]{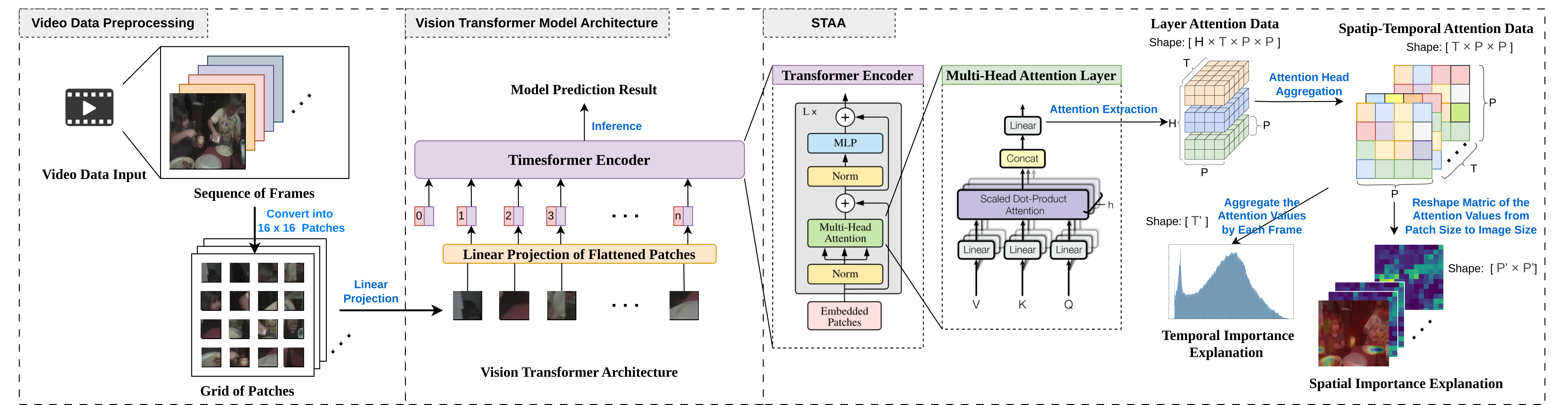}
\caption{Overview of Spatio-temporal Attention Attribution (STAA) method for transformer-based video models}
\label{fig:staa_overview}
\end{figure*}

This section introduces our Transformer attention-based method for analyzing video classification. This approach extracts attention mechanisms in Transformer models to explain both temporal and spatial importance.
Figure \ref{fig:staa_overview} provides a visual overview of STAA method.

\subsubsection{Input and Embedding}
Given an input clip video $X \in \mathbb{R}^{F \times H \times W \times 3}$ consisting of $F$ RGB frames, we decompose each frame into non-overlapping patches of size $P \times P$. Each patch $x^{(p,t)} \in \mathbb{R}^{P^2 \times 3}$ is linearly mapped into an embedding:

\begin{equation}
z^{(0)}_{(p,t)} = Ex^{(p,t)} + e^{pos}_{(p,t)}
\end{equation}
where $E$ is the embedding matrix, and $e^{pos}_{(p,t)} \in \mathbb{R}^D$ represents learnable positional embeddings for spatial position $p$ and temporal position $t$, as marked 0 to N in Figure \ref{fig:staa_overview}. 
Then, $z$ is the overall input to the Transformer, similar to the sequences of words which are used in the language transformer model \cite{vaswani2017attention}.

\subsubsection{Attention Value Extraction}
For each transformer layer $l$ and attention head $a$, we compute query and key vectors for each patch as the Timesformer architecture \cite{TimeSformer}:

\begin{equation}
q^{(l,a)}_{(p,t)} = W^{(l,a)}_Q \text{LN}(z^{(l-1)}_{(p,t)}) \in \mathbb{R}^{D_h}
\end{equation}

\begin{equation}
k^{(l,a)}_{(p,t)} = W^{(l,a)}_K \text{LN}(z^{(l-1)}_{(p,t)}) \in \mathbb{R}^{D_h}
\end{equation}

where: $W^{(l,a)}_Q, W^{(l,a)}_K \in \mathbb{R}^{D_h \times D}$ are learnable weight matrices.  $\text{LN}()$ denotes Layer Normalization. $D_h = D/H$ is the dimension per attention head.

The attention weights for Spatio-Temporal attention are extracted as:
\begin{equation}
\alpha^{(l,a)}_{(p,t)} = \text{SM}\left(\frac{q^{(l,a)}_{(p,t)}}{\sqrt{D_h}}^\top \cdot [k^{(l,a)}_{(0,0)} \{k^{(l,a)}_{(p',t')}\}_{p'=1,...,N}^{t'=1,...,F}]\right) \in \mathbb{R}^{NF+1}
\end{equation}

For space-only attention within each frame:
\begin{equation}
\alpha^{(l,a)\text{space}}_{(p,t)} = \text{SM}\left(\frac{q^{(l,a)}_{(p,t)}}{\sqrt{D_h}}^\top \cdot [k^{(l,a)}_{(0,0)} \{k^{(l,a)}_{(p',t)}\}_{p'=1,...,N}]\right) \in \mathbb{R}^{N+1}
\end{equation}


where: $l$: Index of transformer layer,
$a$: Index of attention head,
$p,p'$: Spatial position indices of patches,
$t,t'$: Temporal frame indices,
$q^{(l,a)}_{(p,t)}, k^{(l,a)}_{(p,t)} \in \mathbb{R}^{D_h}$: Query and key vectors,
$D_h = D/A$: Dimension per attention head,
$N$: Number of patches per frame $(HW/P^2)$,
$F$: Number of frames,
$(0,0)$: Index for classification token,
$\text{SM}()$: Softmax operation.

\begin{algorithm}
    \caption{Spatio-Temporal Attention Attribution (STAA)}
    \label{alg:staa}
    \textbf{Input:} Video clip $X \in \mathbb{R}^{F \times H \times W \times 3}$, Model $\mathcal{M}$\\
    \textbf{Output:} $M_t$, $M_s$
    \begin{algorithmic}
    \STATE Initialize $M_t \in \mathbb{R}^F$, $M_s \in \mathbb{R}^{N \times F}$
    
    \STATE // Extract attention from final layer L
    \STATE $\{\alpha^{(L,a)}\} \leftarrow \mathcal{M}(X)$
    
    \STATE // Aggregate temporal attention
    \STATE $M_t = \frac{1}{AN} \sum_{a=1}^A \sum_{p=1}^N \alpha^{(L,a)}_{(p,:)}$
    
    \STATE // Aggregate spatial attention
    \FOR{$t = 1$ \TO $F$}
        \STATE $M_s[t] = \frac{1}{A} \sum_{a=1}^A \alpha^{(L,a)}_{(:,t)}$
    \ENDFOR
    
    \RETURN $M_t$, $M_s$
    \end{algorithmic}
    \small{\textbf{Notation:} $F$: Number of frames, $N$: Patches per frame $(HW/P^2)$, $A$: Number of attention heads, $L$: Final layer, $\alpha^{(L,a)}$: Attention weights at layer $L$, head $a$, $M_t$: Temporal attention map, $M_s$: Spatial attention maps}
\end{algorithm}

\subsubsection{Temporal and Spatial Attention Aggregation}

To attribute attention to both temporal and spatial dimensions, we aggregate the attention weights across patches and attention heads. 
The attributions can be calculated by processing the attention weights $\alpha$ with Algorithm \ref{alg:staa}. 

The temporal attention map, \( M_t \), provides a frame-level importance score across the video sequence. We achieve this by averaging attention weights over all attention heads and patches within each frame. By doing so, we capture the contribution of each frame to the overall classification decision, revealing which parts of the video sequence are most relevant.
The spatial attention map, \( M_s \), highlights key regions within each frame by focusing on patch-level attention. 
This process allows us to determine which frames and spatial regions within each frame are most influential for the model's classification decision.

\subsubsection{STAA Computational Analysis and Limitations}
STAA achieves significant computational efficiency through its direct utilization of model attention mechanisms. The computational cost can be analyzed in several dimensions:
\texttt{Attention Extraction Mechanism:} STAA extracts feature importance directly from the self-attention weights computed in the Transformer model, requiring only $O(N)$ operations where $N$ is the input video clip size. Specifically, STAA accesses the attention values from the final layer of the Transformer that are naturally computed during video processing. This is fundamentally different from post-hoc methods: SHAP requires $O(2^s · N)$ operations, where $s$ is the number of temporal segments (8 in our implementation), due to its combinatorial sampling approach. LIME's complexity is $O(K · N_p · N)$, where $K$ is the number of selected frames and $N_p$ is the number of perturbation samples (1000 in our implementation). STAA achieves superior efficiency by directly utilizing the model's internal attention mechanism rather than requiring additional sampling or perturbations.

Despite its computational advantages, STAA faces several key limitations:
\texttt{Raw Attention Noise:} When directly visualizing raw attention values as heatmaps overlaid on video frames, we observe significant frame-to-frame fluctuations ("flickering") and diffused attention patterns. The raw attention values often permeate entire frames, making it difficult to identify truly salient regions. This visualization issue suggests the need for attention refinement.
\texttt{Architecture Dependency:} Differ from model-agnostic methods, STAA is specifically designed for Transformer architectures and relies on their self-attention mechanisms. This architectural dependency means STAA cannot be directly applied to other types of video models, such as traditional CNNs.
\texttt{Feature Granularity:} The video Transformers (16×16 pixel patches) actually enables fine-grained spatial analysis and frame-level temporal analysis. This is a significant improvement over post-poc methods. For a standard 224×224 video frame, this creates 196 patches per frame, allowing detailed spatial attention mapping. Each patch can independently receive attention weights. This is sufficient for action recognition tasks. However, applications requiring pixel-precise explanations may need further analysis.

To address the first limitation and further enhance the effectiveness of STAA, we developed a refinement procedure in the following subsection.

\subsection{STAA Enhancement}
\label{sec:Enhancement}
We observed that vanilla STAA resulting heatmaps video exhibits significant frame-to-frame fluctuations with background noise. To address these drawbacks of the vanilla method, we developed an enhancement approach focusing on dynamic thresholding and attention focusing. 

\subsubsection{Dynamic Thresholding for Noise Reduction}
We introduce dynamic threshold $\theta_t$ to frames, calculated as:
\begin{equation}
\theta_t = \mu_t + \lambda \sigma_t
\end{equation}
where $\mu_t$ and $\sigma_t$ are the mean and standard deviation of attention values for frame $t$, respectively. The hyperparameter $\lambda$ controls the threshold's strictness and can be empirically set to between $[0,1]$. In our experiments, it is set to one to allow more attributes to be indicated. If it is set to zero, that means we filter half of the non-important attention attributes as noise.

\subsubsection{Attention Focusing Mechanism}
We apply this threshold to the STAA results:
\begin{equation}
S'[t,p] = \begin{cases}
S[t,p] & \text{if } S[t,p] \geq \theta_t \\[2pt]
0 & \text{otherwise}
\end{cases}
\end{equation}
This operation effectively focuses attention on the most salient spatio-temporal regions, reducing noise and improving the clarity of our explanations.

\subsubsection{Attention Visualization}
\begin{figure}[h]
\centering
\includegraphics[width=0.5\textwidth]{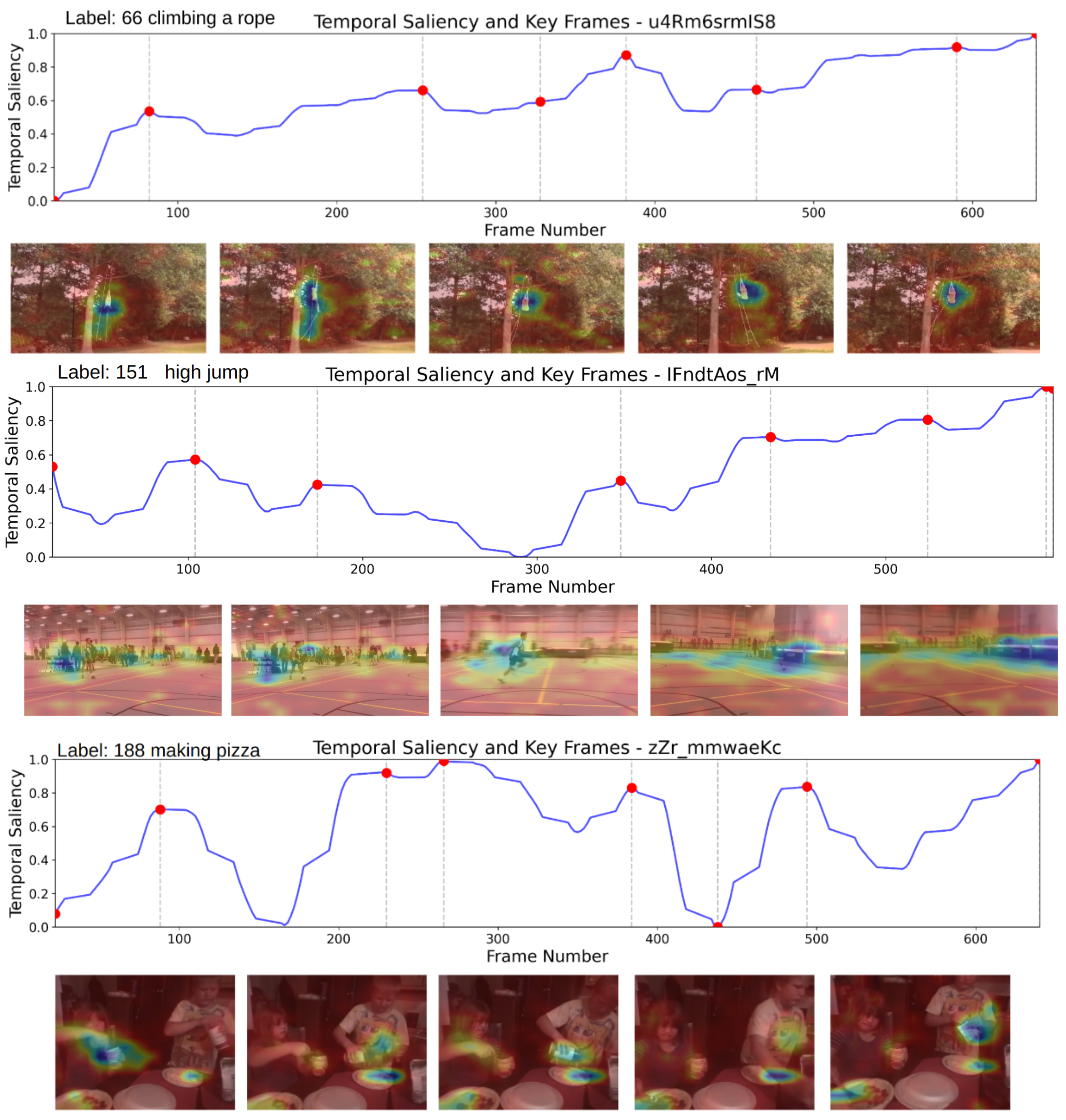}
\caption{Example of attention visualization: heatmap overlaid on a video frame, where blue colors indicate regions of higher importance, while red indicate lower importance.}
\label{fig:heatmap_example}
\end{figure}

To enable intuitive human interpretation of the model's focus and facilitate qualitative analysis of attention patterns, we transform the attention values into visualizable heatmaps.
First, we normalize the filtered attention values to [0,1] range. Then, we blend the colorized heatmap with the original frame:
\begin{equation}
H_t = \frac{S'_t - \min(S'_t)}{\max(S'_t) - \min(S'_t)}
\end{equation}
where $H_t$ is the normalized attention map. The resulting visualization overlays attention heatmaps on video frames. 
Here are result examples shown in Figure \ref{fig:heatmap_example} with a color spectrum from red to blue.

This visualization highlights the regions and frames that most significantly influence its predictions. Such insights can be used for model evaluation.
The complete implementation, including heatmap video explanation samples and source code, is available in our GitHub repository\footnote{\url{https://github.com/ZeruiW/VideoXAI}}.

In the subsequent sections, we present extensive experimental results that demonstrate the effectiveness of both the vanilla and enhanced versions of our STAA method. We also provide comparisons with SHAP and LIME approaches, highlighting the unique contributions offered by STAA to the field of explainable AI in video analysis, as well as showcasing how our enhancements address the initial limitations of the method.

\begin{mdframed}[skipabove=5pt,innertopmargin=5pt,linecolor=black,roundcorner=5pt,backgroundcolor=gray!25] 
    Answer to RQ1: STAA, as a new XAI method, extracts and processes attention values that capture both spatial importance within frames and temporal relevance across the video. 
    \end{mdframed}

\section{Experimental Evaluation and Comparative Analysis}

We conducted comprehensive experiments to evaluate the effectiveness of STAA against established XAI methods using a large-scale video understanding dataset. Our experimental design addresses three key aspects: dataset selection and preprocessing, model architecture configuration, and evaluation methodology.
We aim to evaluate Faithfulness and Monotonicity metrics for XAI. Faithfulness indicates how accurately each method's explanations reflect the model's decision-making process. Monotonicity indicates the consistency of feature importance rankings provided by each method.

\subsection{Experimental Design and Setup}

The experiments were conducted using the Kinetics-400 dataset \cite{kay2017kinetics}, a comprehensive benchmark containing approximately 240,000 video clips across 400 human action classes. This dataset represents diverse spatio-temporal patterns, from simple actions to complex sequences. Its widespread adoption in the video AI community \cite{TimeSformer,vivit,videoswin,selva2023video,carreira2017quo} enables meaningful comparisons with existing work while providing sufficient complexity to assess XAI performance across varied scenarios. To address potential class imbalance effects on XAI evaluation, we implemented a balanced sampling strategy across action classes. We randomly selecting thirty video file from each of the four-hundreds action classes, resulting in a total of twelve-thousand videos for analysis. 

The TimeSformer model \cite{TimeSformer} was employed as the model architecture, having achieved an accuracy of 78.0\% on the Kinetics-400 validation set. This choice was motivated by TimeSformer's state-of-the-art performance and its representative implementation of self-attention mechanisms. The model processes input videos at 224×224 pixel resolution.

The experiments were conducted on a system equipped with an NVIDIA GeForce RTX 4090 GPU, running Ubuntu 22.04.5 LTS. This configuration enables efficient processing of video data. All implementation code and experimental configurations are available in our public repository to facilitate reproducibility and further research in video XAI methods.

\subsection{Quantitative XAI Evaluation Metrics}

To assess the explanations generated by our XAI methods, we developed two metrics: \texttt{faithfulness} and \texttt{monotonicity}. These metrics are designed according to the fundamental properties of feature importance. 

The concept of faithfulness in XAI evaluation was initially developed for image classification tasks \cite{faithfulness,Wang2024tcc}, measuring how accurately an explanation reflects a model's decision-making process. We extend this concept to video XAI analysis.
It quantifies the degree to which the identified important features truly influence the model's output.
Monotonicity serves as a more detailed complementary metric.
Monotonicity evaluates whether features assigned higher importance scores indeed have a more significant impact on the model's predictions.

\subsubsection{Faithfulness}

Faithfulness quantifies how accurately an explanation represents the model's decision-making process. We define this metric as:

\begin{equation}
\text{faithfulness} = 1 - \frac{1}{N} \sum_{i=1}^N |f_{\text{norm}}(x_i) - f_{\text{norm}}(x_i \setminus s_i)|
\end{equation}

where $f_{\text{norm}}(x_i)$ is the normalized prediction for the original input $x_i$, $f_{\text{norm}}(x_i \setminus s_i)$ is the normalized prediction after masking the features $s_i$ identified as important, and $N$ is the number of test samples.

In our implementation, We use a mask ratio of 0.7, removing 70\% of the regions deemed most important by the explanation.
We apply min-max normalization to the predictions before calculating the difference.
The faithfulness score ranges from 0 to 1, with higher values indicating better XAI performance. A score closer to 1 suggests that masking important features significantly changes the model's normalized prediction, indicating that the explanation has accurately identified crucial features.

\subsubsection{Monotonicity}

Monotonicity assesses whether the importance rankings produced by an explanation method are consistent with their impact on the model's predictions. We quantify this using Kendall's rank correlation coefficient (Kendall's tau) \cite{kendall1938new}, a statistical measure that evaluates the ordinal association between two rankings:

\begin{equation}
\text{monotonicity} = \tau({p_k}, {d_k})
\end{equation}
where:
$p_k$ represents $K$ numbers ascending sequence of masking ratios, for example, [0.1, 0.2, ..., 0.9] in ten ratios,
$d_k = f_{\text{softmax}}(x)[c] - f_{\text{softmax}}(x \setminus s_k)[c]$ represents the changes in prediction probability when masking top-$K$ important features
$s_k$ represents regions masked according to their importance, 
$\tau$ is the rank correlation coefficient \cite{kendall1938new}.

In our context, a high positive $\tau$ value suggests that as we mask more important features (increasing $p_k$), we observe proportionally larger changes in the model's predictions ($d_k$), confirming that the XAI method has consistently identified the important features. Conversely, a value close to -1 indicates that the rankings are completely opposite, suggesting that features identified as important by the XAI method actually have a worse impact on the model's predictions.

This metric can be applied on either temporal feature or spatial feature or even both. For temporal analysis, it evaluates whether masking important time segments leads to consistent prediction changes. For spatial analysis, it assesses the impact of masking important regions within frames. When applied to Spatio-Temporal features, we mask them both. It measures the overall consistency of the selected XAI methods.

\subsection{Results Analysis and Method Comparison}

Table \ref{tab:xai_fidelity} presents the results of our evaluation using faithfulness and monotonicity metrics on the Kinetics-400 dataset. 

\begin{table}[ht]
    \renewcommand{\arraystretch}{1.2}
    \caption{XAI Evaluation Results}
    \label{tab:xai_fidelity}
    \centering
    \setlength{\tabcolsep}{4pt}
    \begin{tabular}{@{}lccc@{}}
    \hline
    \textbf{Method} & \textbf{Faithfulness} & \textbf{Monotonicity} & \textbf{Computation Time (s)} \\
    \hline
    SHAP & 0.769 ± 0.122 &  -0.535 ± 0.108 & 5.59 ± 0.44  \\
    LIME & 0.525 ± 0.388 & -0.411 ± 0.186 & 46.83 ± 8.68 \\
    STAA (Vanilla) & 0.190 ± 0.373 & 0.496 ± 0.437 & \textbf{0.16 ± 0.01} \\
    STAA (Enhanced) & \textbf{0.844 ± 0.116} & \textbf{0.850 ± 0.030} & \textbf{0.16 ± 0.01} \\
    \hline
    \end{tabular}
\end{table}

The results demonstrate varying performance across different XAI methods. Let's analyze each method's performance:
SHAP shows moderate faithfulness (0.769 ± 0.122) but poor monotonicity (-0.535 ± 0.108). 
LIME exhibits low faithfulness (0.525 ± 0.388) and monotonicity (-0.411 ± 0.186), indicating that LIME's explanations may not accurately reflect the model's decision process and lack consistency in feature importance rankings.
STAA (Vanilla), using raw attention values, shows poor faithfulness (0.190 ± 0.373) but small positive monotonicity (0.496 ± 0.437). This suggests that the noise in attention values significantly impact the accuracy of XAI evaluation.
STAA (Enhanced), our enhanced method demonstrates superior performance in both metrics. It achieves the highest faithfulness (0.844 ± 0.116) and monotonicity (0.850 ± 0.030) scores among all methods.

The improvement of STAA from Vanilla to Enhanced is substantial, indicating the effectiveness of our threshold approach. The high faithfulness score suggests that STAA (Enhanced) provides explanations that accurately reflect the model's decision-making process. The high monotonicity score indicates that the importance rankings produced by our method are highly consistent and reliable across different levels of feature masking.

Both Vanilla and Enhanced STAA methods demonstrate a significant computational advantage over SHAP and LIME, with an average runtime of just 0.16 ± 0.01 seconds. The enhancement algorithm presented in Section \ref{sec:Enhancement} adds negligible overhead to the overall processing, especially relative to the deep-learning model's complexity, rendering this millisecond-level cost effectively inconsequential. This high efficiency, combined with superior performance metrics, makes STAA (Enhanced) exceptionally well-suited for explaining video analysis tasks. More detailed results, visualization and reproducible code can be found in our GitHub repository. 

Overall, the findings emphasize the strengths of our Spatio-Temporal Attention Attribution approach, particularly in its enhanced form. Combining high faithfulness and monotonicity with minimal computation time, STAA (Enhanced) offers an efficient method for interpreting complex video analysis tasks. 

\begin{mdframed}[skipabove=10pt,innertopmargin=10pt,linecolor=black,roundcorner=5pt,backgroundcolor=gray!25] 
    Answer to RQ2: Our experimental results demonstrate that STAA significantly outperforms traditional model-agnostic methods including SHAP and LIME. STAA achieves superior faithfulness (0.844 ± 0.116) and monotonicity (0.850 ± 0.030) scores, while reducing computation time by more than 97\%.
\end{mdframed}

\section{Real-Time Video XAI Cloud Architecture}
This section presents a cloud-based architecture that enables real-time explanation generation for video analysis, addressing the critical challenge of providing sub-hundreds-milliseconds latency explanations for edge computing applications \cite{Milliseconds}. We first establish the architectural framework necessary for real-time video XAI, then evaluate its performance against industry-standard latency requirements.

\subsection{Architecture Components Overview}

To achieve real-time video XAI processing while maintaining system efficiency, we propose a distributed architecture that uses cloud server's computing resources while minimizing edge device computational overhead. The architecture employs cloud-based GPUs for compute-intensive tasks, including model inference and XAI processing, while edge devices handle data acquisition and visualization, enabling real-time processing within the established hundreds-milliseconds latency threshold \cite{Milliseconds}.

\begin{figure}[h]
\centering
\includegraphics[width=0.8\linewidth]{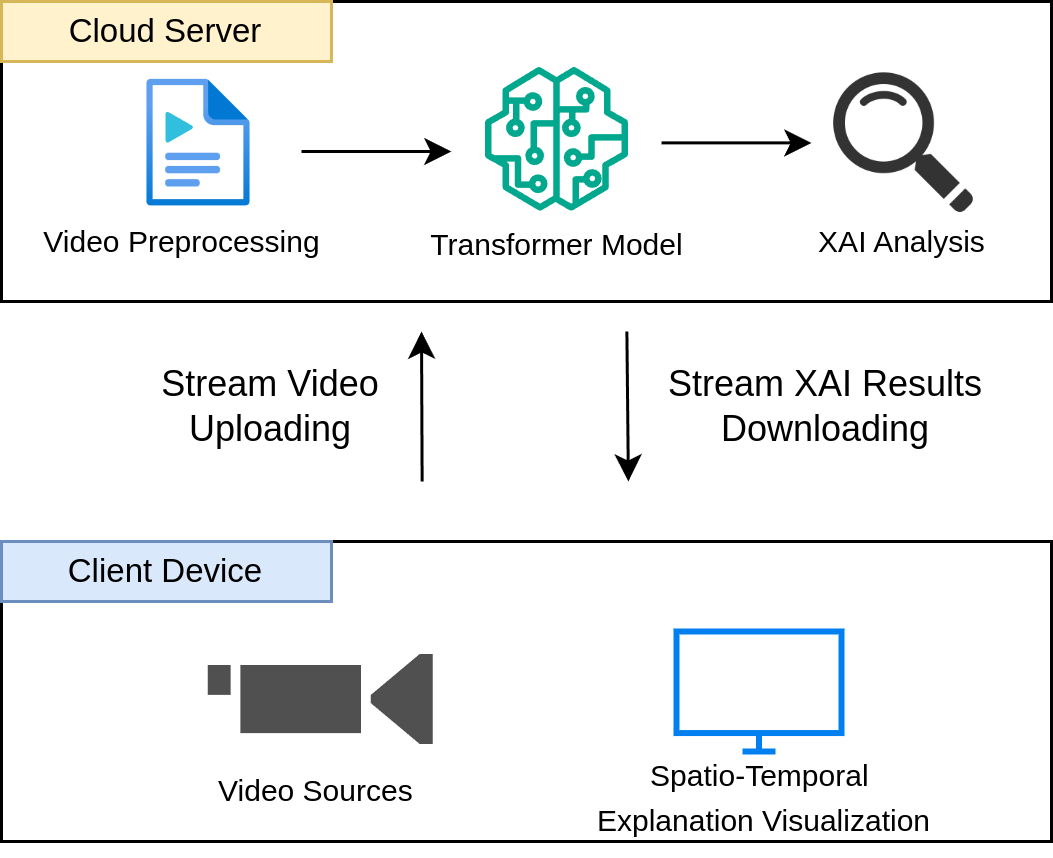}
\caption{Cloud-based architecture for real-time video XAI}
\label{fig:system_architecture}
\end{figure}

Our proposed architecture integrates five main components to enable efficient real-time video XAI analysis: the \texttt{Video Sources} component serves as the data input, providing various streams such as live camera feeds; the \texttt{Video Preprocessing} module performs essential tasks including temporal segmentation, frame extraction, and resizing; the \texttt{Transformer Model} implements state-of-the-art video analysis capabilities such as TimeSformer \cite{TimeSformer}; working in tandem with this model is the \texttt{XAI Analysis} component that generates feature attribution explanation through the adopted XAI method; and finally, the \texttt{Spatio-Temporal Explanation Visualization} component renders these XAI outputs in human-interpretable formats, for example, heatmap video. These components are distributed between cloud and client devices, as illustrated in Fig. \ref{fig:system_architecture}. The cloud service handles the compute-intensive components, while client-side devices focus on data collection and real-time visualization of the explanations. This design specifically addresses the challenges faced by resource-constrained edge devices for complex XAI tasks \cite{edgecloudecosystem,wang2024xaiport}.

\subsection{Performance Evaluation for Real-Time Scenarios}

To evaluate the applicability of our STAA method, we implemented a test for real-time video XAI. This system demonstrates the practical feasibility of generating explanations for video streams.

\subsubsection{Server development}
Our implementation consists of two main components.

\texttt{Client}: The client reads video files using the PyAV library, which allows for efficient frame extraction. It segments the video into batches which suit for model inputs, and sends these batches to the server using ZeroMQ \cite{ZeroMQ}, a high-performance asynchronous messaging library. The client side enables real-time streaming of video data to the server side.

\texttt{Server}: The server, built around the AI model and XAI methods, receives frame batches and processes them in real-time. 
Video AI model inference the frame batches. XAI method generates heatmaps, and produces explanations for each batch of frames. The server utilizes multi-threading to separate frame reception from processing, ensuring continuous data flow.

\subsubsection{Real-time Processing Strategies}

Our system achieves real-time processing through several key strategies:

\texttt{Fine Batch Processing}: The server processes video in batches of frames, aligning with the Timesformer model's input requirements. This approach balances the trade-off between processing granularity and computational efficiency. The smaller batches could provide finer-grained XAI analysis, while the larger batches means process longer video in each model inference. In the test, we use eight-frame batch size represents the minimal unit that the model can process.

\texttt{STAA Method}: The system employ our STAA method to extract and process attention values from the model's internal mechanisms. It generates spatio-temporal feature importance through the following steps: (1) decomposing frames into non-overlapping patches, (2) computing attention weights $\alpha^{(l,a)}_{(p,t)}$ from the final transformer layer, and (3) aggregating these weights across attention heads to produce temporal importance scores $M_t$ and spatial importance maps $M_s$. The method employs dynamic thresholding ($\theta_t = \mu_t + \lambda \sigma_t$) to reduce noise and enhance visualization clarity. 

\texttt{Asynchronous Frame Processing}: The system implements an asynchronous pipeline where frame extraction, model inference, and explanation generation operate independently. This design prevents bottlenecks and ensures continuous processing flow, essential for real-time performance.

\subsubsection{Performance Evaluation and Analysis}
\begin{figure}[h]
\centering
\includegraphics[width=1.0\linewidth]{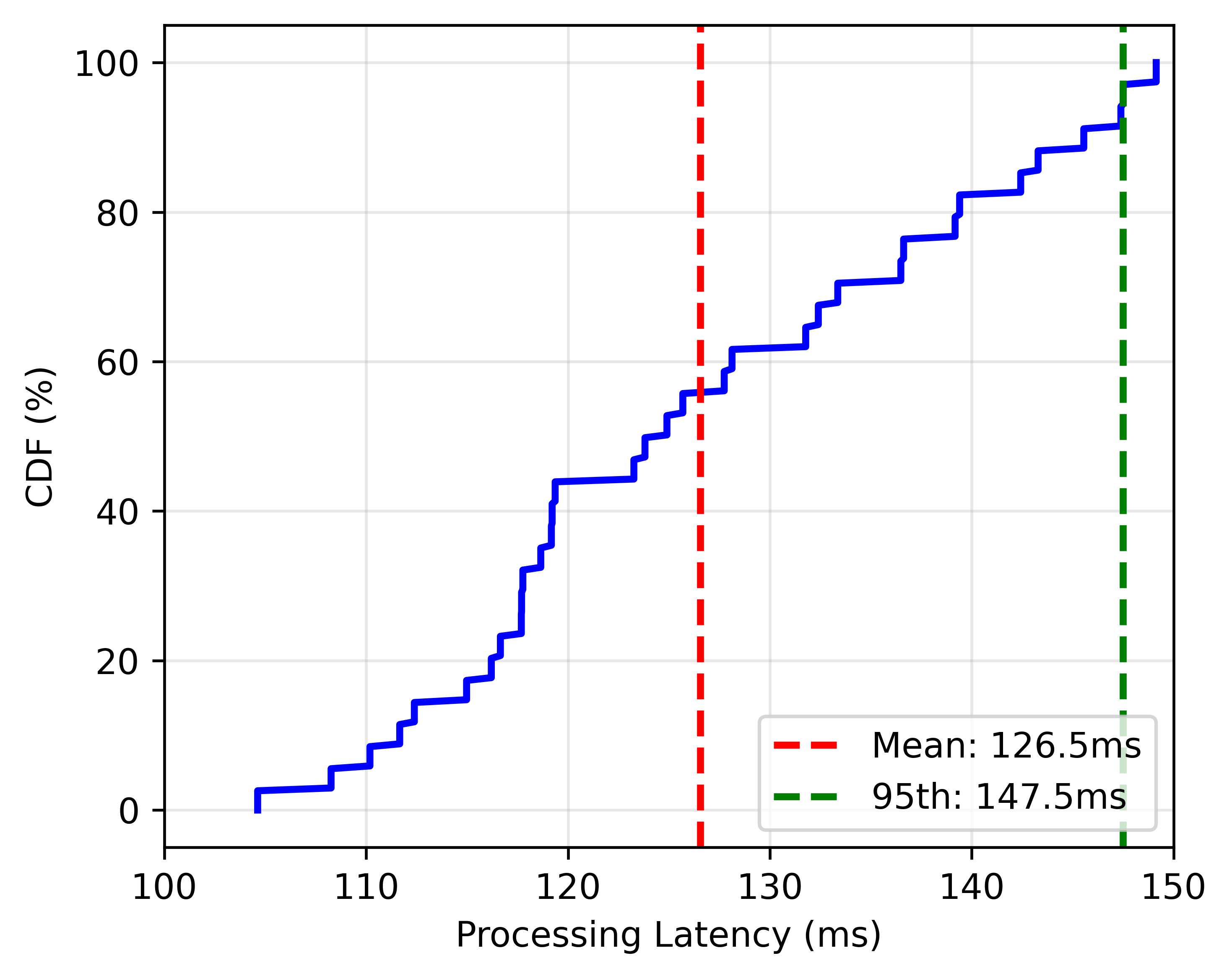}
\caption{Cumulative Distribution Function (CDF) of the STAA processing latency}
\label{fig:latency}
\end{figure}

We conducted performance testing of our real-time video XAI system, focusing on the \texttt{Latency} metrics.
Measured as the end-to-end provess time between receiving a batch of frames and completing feature attribution explanation. The test results are illustrated in a Cumulative Distribution Function diagram in Figure \ref{fig:latency}.

These results demonstrate that our system can process frames between a hundred milliseconds to a hundred and fifty milliseconds, allowing for near-immediate explanation generation. Besides, the general action classification tasks not require frame by frame explanation. Therefore, in practical applications, practitioners can adjusting the sampling rate according to their specific needs. For instance, since most human actions evolve over multiple frames, sampling eight to thirty-two frames per video clip (around ten seconds) is typically sufficient for accurate action recognition and explanation generation \cite{TimeSformer,vivit}.

Traditional post-hoc XAI methods such as SHAP or LIME, require multiple passes over the entire video. The latency for these methods (5.59 seconds for SHAP and 46.83 seconds for LIME) are significantly longer than the required time for real-time application. 
Our STAA-based system generates explanations in a single pass of model inference. This fundamental difference in approach enables our system to provide explanations with minimal delay, making it suitable for real-time applications.


To facilitate reproducibility and future development, we provide the source code for real-time video XAI server implementation on GitHub \footnote{https://github.com/ZeruiW/RealTimeServer}. This codebase serves as a practical demonstration of our method and can be used as a foundation for developing more advanced XAI services in video analysis. 
The repository provides a guide for setting up the server and client.

In conclusion, our implementation demonstrates the practical feasibility of real-time video XAI, opening up new possibilities for applications requiring immediate insights into AI decision-making processes in video analysis tasks.

\begin{mdframed}[skipabove=10pt,innertopmargin=10pt,linecolor=black,roundcorner=5pt,backgroundcolor=gray!25] 
    Answer to RQ3: We designed and implemented a cloud-based architecture for real-time video XAI, employing GPUs on the cloud server for efficient model inference and XAI processing. This approach enables client-side devices to achieves an average latency of hundreds milliseconds per frame explannation, demonstrating the feasibility of real-time video XAI for edge devices.
    \end{mdframed}

\section{Conclusion}
This paper presents an advancement in the field of explainable AI for video analysis, addressing the need for XAI in increasing video AI deployment. Our proposed STAA method represents a new approach to generating explanations for video Transformer models, overcoming the limitations of existing post-hoc explanation techniques.
The key contributions of this work are threefold:
(1). We introduce STAA, a method that provides spatio-temporal explanations simultaneously. The computational cost is reduced by 97\% compared to SHAP and LIME. 
(2). Through extensive experimental evaluation on the Kinetics-400 dataset, we demonstrate that STAA significantly outperforms established methods, leading to higher faithfulness and monotonicity scores.
(3). We present a cloud-based architecture for real-time video XAI, showcasing STAA's practical applicability in scenarios requiring immediate explanations. Our implementation measured an average latency of less than a hundred-fifty milliseconds per frame, demonstrating the feasibility of real-time explainable video analysis.

In the future, we will further examine the resilience of the STAA method under various adversarial attacks and test more adaptability across different novel Transformer video models.




\bibliographystyle{IEEEtran}
\bibliography{Reference}

\vspace{-2cm}
\begin{IEEEbiography}[{\includegraphics[width=1in,height=1.25in,clip]{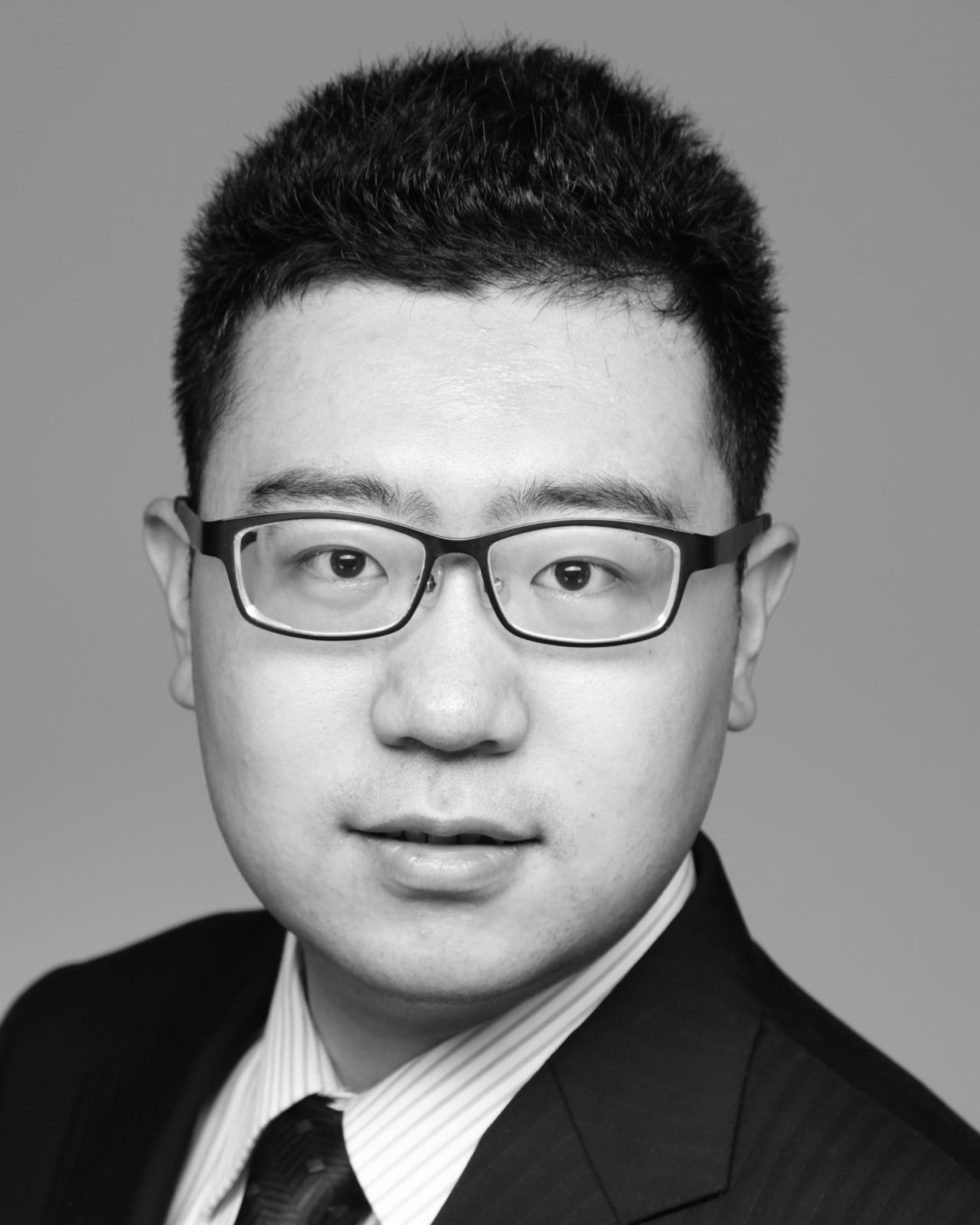}}]{Zerui Wang}is a Ph.D. candidate in the Department of Electrical and Computer Engineering at Concordia University. His research interests include Explainable AI (XAI), cloud AI services, and applied AI. Zerui's work has been published in IEEE Transactions on Cloud Computing and international conferences such as ICSE. He focuses on enhancing explainable AI in cloud computing environments. Contact him at zerui.wang@concordia.ca.
\end{IEEEbiography}
\vspace{-15cm}
\begin{IEEEbiography}[{\includegraphics[width=1in,height=1.25in,clip,keepaspectratio]{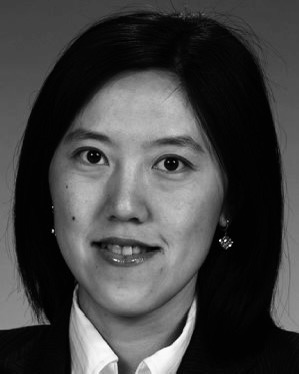}}]{Dr. Yan Liu} is a Full Professor and Gina Cody Research and Innovation Fellow at Concordia University. Before the faculty position, Yan worked as a Senior Research Scientist at the National ICT Australia (NICTA) laboratory and US Department of Energy Pacific Northwest National Laboratory with ten years of experience with large-scale software systems. As a tenured faculty, Yan's research is generously funded by NSERC Discovery Grants, Quebec FRQNT New Research Award, and MITACS and industry collaborators in the domains of telecommunication, health care, senor networks, NLP for public services, cloud game servers, and digitization of building architecture design. Yan has two US patents granted. Her recent work is defining an evaluation framework for explanation consistency. Contact her at yan.liu@concordia.ca
\end{IEEEbiography}

\end{document}